%% file: main.tex
\PassOptionsToPackage{dvipsnames,table}{xcolor}
\documentclass[10pt,twocolumn,letterpaper]{article}

\usepackage{cvpr}              % To produce the CAMERA-READY version
\usepackage{times}
\usepackage{epsfig}
\usepackage{graphicx}
\usepackage{amsmath}
\usepackage{amssymb}
\usepackage{booktabs}
\usepackage{multirow}
\usepackage{multicol}
\usepackage{diagbox}
\usepackage{utfsym}
\usepackage{subcaption}
\usepackage{stfloats}
\usepackage{float}
\usepackage{algorithmic}
\usepackage{algorithm}
\usepackage{listings}
\usepackage[titletoc,title]{appendix}
\usepackage[accsupp]{axessibility}
\usepackage[inkscapelatex=false]{svg}
\usepackage[font=footnotesize]{caption}
\input{preamble}

\definecolor{cvprblue}{rgb}{0.21,0.49,0.74}
\usepackage[pagebackref,breaklinks,colorlinks,citecolor=cvprblue]{hyperref}
\usepackage[misc]{ifsym}
\newcommand\blfootnote[1]{%
\begingroup
\renewcommand\thefootnote{}\footnote{#1}%
\addtocounter{footnote}{-1}%
\endgroup
}

\def\paperID{11483} % *** Enter the Paper ID here
\def\confName{CVPR}
\def\confYear{2024}

\title{MV-Adapter: Multimodal Video Transfer Learning for Video Text Retrieval}

\author{Xiaojie Jin$^{1*}$\textsuperscript{\Letter}, Bowen Zhang$^{1*}$, Weibo Gong$^1$, Kai Xu$^1$, Xueqing Deng$^1$, \and Peng Wang$^1$, Zhao Zhang$^2$, Xiaohui Shen$^1$, Jiashi Feng$^1$\\
$^1$Bytedance Inc., $^2$Hefei University of Technology\\
{\tt\small \{jinxiaojie,zhangbowen.17,gongweibo,xukai.1993,xueqingdeng\}@bytedance.com}\\
{\tt\small \{peng.wang,shenxiaohui.kevin,jshfeng\}@bytedance.com} \ \ {\tt\small cszzhang@gmail.com}
}

\begin{document}
\maketitle
\blfootnote{$^{*}$Equal contribution. Bowen Zhang did the work during an internship.\newline \hspace*{1.9em}{\scriptsize \Letter} Corresponding author: Xiaojie Jin.}

\input{sec/0_abstract}    
\input{sec/1_intro}
\input{sec/2_related_work}
\input{sec/3_method}
\input{sec/4_experiment}
\input{sec/5_result}
\input{sec/6_conclustion}
{
    \small
    \bibliographystyle{ieeenat_fullname}
    \bibliography{main}
}

\begin{appendices}
\input{sec/X_suppl}

\end{appendices}
\end{document}

%% file: preamble.tex
\usepackage[dvipsnames]{xcolor}

%% file: sec/0_abstract.tex
\begin{abstract}
   State-of-the-art video-text retrieval (VTR) methods typically involve fully fine-tuning a pre-trained model (e.g. CLIP) on specific datasets. However, this can result in significant storage costs in practical applications as a separate model per task must be stored. To address this issue, we present our pioneering work that enables parameter-efficient VTR using a pre-trained model, with only a small number of tunable parameters during training.  Towards this goal, we propose a new method dubbed \textbf{M}ultimodal \textbf{V}ideo Adapter (MV-Adapter) for efficiently transferring the knowledge in the pre-trained CLIP from image-text to video-text. Specifically, MV-Adapter utilizes bottleneck structures in both video and text branches, along with two novel components. The first is a \emph{Temporal Adaptation Module} that is incorporated in the video branch to introduce global and local temporal contexts. We also train weights calibrations to adjust to dynamic variations across frames. The second is \emph{Cross Modality Tying} that generates weights for video/text branches through sharing cross modality factors, for better aligning between modalities. Thanks to above innovations, MV-Adapter can achieve comparable or better performance than standard full fine-tuning with negligible parameters overhead. Notably, MV-Adapter consistently outperforms various competing methods in V2T/T2V tasks with large margins on five widely used VTR benchmarks (MSR-VTT, MSVD, LSMDC, DiDemo, and ActivityNet). Codes will be available on \href{https://github.com/zhangbw17/MV-Adapter}{github}. 
\end{abstract}

%% file: sec/1_intro.tex
\vspace{-5mm}\section{Introduction}
\label{sec:intro}
\begin{figure}
\centering
\includegraphics[width=0.66\linewidth]{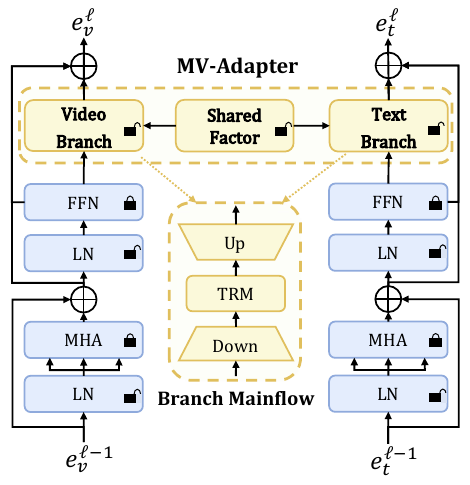}
\vspace{-4mm}
\caption{The overall pipeline of MV-adapter with the illustration of the basic structure of video/text branches. Only a small part of model is tunable during training, highlighted by the ``unlock" symbol.}
\label{fig:pipeline_overall}
\vspace{-6mm}
\end{figure}
Video text retrieval (VTR) \cite{lei2021less,o1,o2,o3,o4,o5,c4c,twi,hy,xclip,straight,xpool}, aiming to obtain the rankings of videos/texts in a repository given text/video queries (i.e. T2V and V2T respectively) has a wide range of practical applications. Recently, with the surge of large-scale pre-trained image-text models, particularly CLIP~\cite{CLIP}, transferring the knowledge learned in CLIP to VTR tasks by fully fine-tuning has become the de-facto paradigm and adopted by state-of-the-art methods \cite{c4c,twi,hy,xclip,straight,xpool}.

However, these methods suffer from substantial storage overhead in practical applications as each task necessitates the storage of a distinct model. This issue becomes more severe when the pre-trained model size increases and multiple VTR tasks need to be solved, thus hindering the application in real-world scenarios. For instance, in mobile apps, model size is restricted to reduce downloading time and/or app package size, as a larger size may decrease active users. In cloud services, many models need to be trained to solve functionally similar but customized tasks, making standard fine-tuning unfavorable in such cases. Additionally, since most VTR tasks have relatively small training sets, performing full fine-tuning on these datasets results in instability and poor performance, as demonstrated in ~\cite{peters2019tune,dodge2020fine}.

To solve this problem, we introduce a new task to perform Parameter Efficient transfer learning of VTR (PE-VTR), i.e. only a small number of parameters are tunable during training while the majority weights are frozen. This results in high degree of parameter sharing between models, with each model requiring only a small number of additional parameters for a new task.  Generally speaking, there are two challenges in tackling PE-VTR: (i) adapt image-text pre-trained models to video-text and sufficiently learn the temporal context and cross-modal correlations. (ii) ensure that the model is parameter-efficient with negligible parameter overhead while maintaining performance. Initially, we revisit current methods \cite{c4c,hy} that perform full fine-tuning by adapting them to PE-VTR and freezing their CLIP backbone. However, to our surprise, they all fall far behind the full fine-tuning counterparts in terms of performance (\cf  \cref{subsec:result}). In addition to inferior performance, these methods also have other limitations that make them unsuitable for PE-VTR. Some \cite{adaptmlp,convpass,framework} are designed only for a single modality (image or text) and ignore the temporal modeling and/or the interactions between multimodal features. Others introduce significant parameter overhead, which contradicts the purpose of PE-VTR \cite{st}. The above analysis demonstrates that there is still a significant research gap in addressing PE-VTR.

In this paper, we propose a novel method called multimodal video adapter for tackling PE-VTR. As illustrated in \cref{fig:pipeline_overall}, MV-Adapter has two branches for video and text respectively. Each branch uses a bottleneck-style structure with three main operations: \texttt{Downsample-Transformer-Upsample}. We propose two novel components to address the challenges of PE-VTR. First, we introduce a temporal adaptation (TA) module in the video branch to enhance the temporal modeling capability. Unlike previous video adapters that apply identical weights across frames, we generate dynamic weights from both global and local features to better capture temporal variations in videos. Second, we propose a Cross Modality Tying (CMT) module that generates weights for the video and text branches from a modality shared parameter space. By implicitly "shortcutting" weights, models are apt to learn semantically aligned features between modalities, which aligns with the objective of VTR to bring multimodal representations closer together. 
\begin{figure}
\centering
\includegraphics[width=\linewidth]{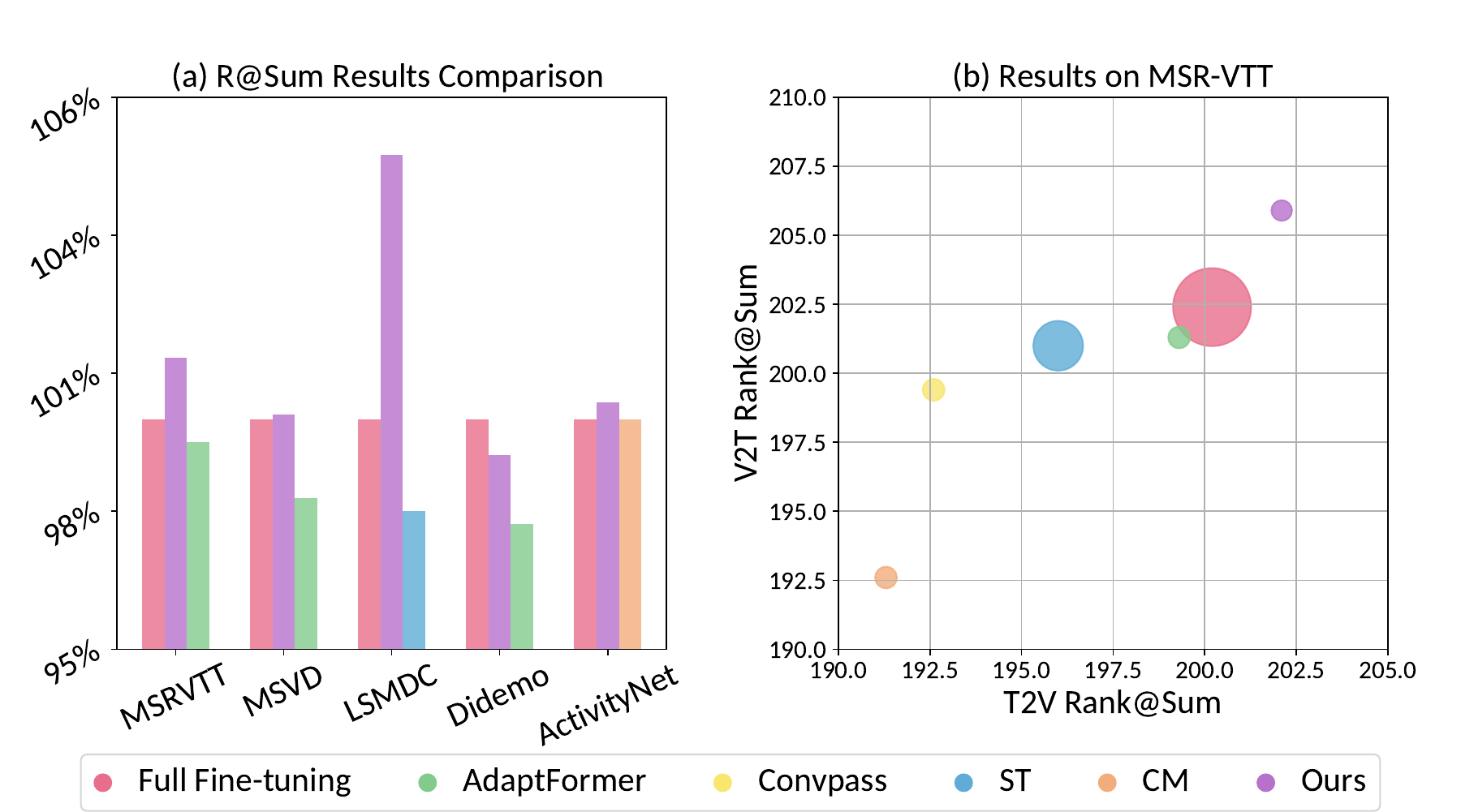}
\vspace{-6mm}
\caption{(a) The overall results on five widely used VTR benchmarks, We present the R@Sum (sum of the R@1, R@5, and R@10) results for the Text-to-Video and Video-to-Text tasks for full fine-tuning, ours, and the best baseline method, displayed as a ratio to the R@Sum of full fine-tuning. (b) Comparison of Text-to-Video and Video-to-Text R@Sum for different methods on MSR-VTT, where the radius of the circle is positively correlated with the trainable parameters.}
\label{fig:overall_result}
\vspace{-6mm}
\end{figure}
Equipped with the above innovations, MV-Adapter is both parameter-efficient and performant on the PE-VTR task. Through extensive experiments on the five most commonly used VTR benchmarks, MV-Adapter achieves comparable or better results than the fully fine-tuned model, with negligible overhead (only 2.4\% extra parameters). Compared with other VTR methods and Adapters, MV-Adapter significantly surpasses its competitors with large margins in V2T/T2V performance while using fewer parameters.

In summary, the contributions of our method are:
\begin{itemize}
    \item We are among the first to take on the task of parameter-efficient VTR (PE-VTR) to promote its real-world applications. Current VTR methods face issues with applicability due to their large parameter overhead. 
    \item We propose a novel method called MV-Adapter to tackle PE-VTR, including two novel modules: temporal adaptation and cross modality tying. These modules effectively address the adaptation and efficiency challenges.
    \item We conduct extensive experiments on five widely-used VTR datasets to evaluate the performance of MV-Adapter. The results show that it achieves comparable or even better performance than standard full fine-tuning, and it also achieves the best trade-off between performance and efficiency among all competing methods. 
\end{itemize}

%% file: sec/2_related_work.tex
\section {Related Work}
\subsection{Image-text Pre-trained Model}
With the increasing demand for model capabilities, a lot of works \cite{CLIP,florence,mahajan2018exploring,kolesnikov2019large,jia2021scaling,flamingo} leverage large-scale Internet data to learn general representations. These works outperform previous methods on numerous downstream tasks, demonstrating the effectiveness of self-supervised learning on big data. Among many others, \cite{CLIP} is widely used by previous methods as backbone model. It has been demonstrated to provide solid prior knowledge for the downstream tasks, which is a better initialization than training from scratch. \cite{CLIP,florence,flamingo,jia2021scaling} show encouraging results using the paradigm of pre-training followed by transfer learning.

\subsection{Parameter-Efficient Transfer Learning}
\label{sec:PETL-related_work}
As the model grows larger, fully fine-tuning all parameters is prohibitively costly. Therefore, the demand for parameter-efficient transfer learning (PETL) increases. PETL methods can be broadly classified into two categories. The first category updates partial parameters in the model sparsely~\cite{bitfit,diff}. The second approach update only newly added parameters/modules. For example, \cite{lora,prefix} add or modify the $QKV$ matrix in the transformer module. \cite{vpt,ppt} add learnable parameters to the input in the form of prompts. Adapter~\cite{framework} is one of the mainstream PETL methods in this direction. Early works \cite{rebuffi2017learning,rebuffi2018efficient} introduce adapters to Computer Vision. \cite{adaptmlp} proposes a simple adapter AdaptFormer based on ViTs~\cite{vit}. The Convpass\cite{convpass} and ST-Adapter~\cite{st} utilize the spatial invariance and the temporal information of videos respectively. Adapters are also widely used in NLP~\cite{adapt,framework}.

Adapter for multimodal tasks is relatively scarce. Previous works \cite{adaptmlp,convpass,framework} focus on unimodal tasks like classification. VL-adapter \cite{vl} only adapts the text stream, while the visual projection of CLIP is fine-tuned. \cite{cm_thu} adjusts CLIP by inserting a few parameterization
layers while ignoring the temporal modeling in transfer learning. UniAdapter \cite{lu2023uniadapter} and Aurora \cite{wang2023parameter} respectively transfer BLIP \cite{li2022blip} to some multimodal tasks in a parameter efficient manner through knowledge sharing and mode approximation. Different from above works, our method takes both modalities into consideration and adapts to the temporal domains.

\subsection{Video Text Retrieval}
\cite{msrvtt,msvd,anet,didemo,lsmdc,ycc2} are most widely used datasets in video-text retrieval (VTR). Early works \cite{o1,o2,o3,o4,o5} use offline features extracted by expert models for modal fusion. Since the emergence of the CLIP~\cite{CLIP}, \cite{c4c,straight} transfer CLIP to VTR task. They show CLIP significantly outperformed the previous models. Afterward, using CLIP for the video-text retrieval task became a new paradigm. \cite{xpool} uses text features as query vectors and applies the attention mechanism to image features. \cite{twi} designs a fine-grained token-wise interaction to calculate the similarity score. \cite{hy} designs a hierarchical aggregation mechanism of features. \cite{xclip} designs a multi-grained interaction mechanism. However, all these works fine-tune the entire parameter set of CLIP, thus incurring high storage overhead. We focus on the parameter-efficient learning of VTR.

%% file: sec/3_method.tex
\definecolor{commentgreen}{RGB}{0,128,0}
\definecolor{strorange}{RGB}{205, 120, 50}
\section {Methodology}
\begin{figure}
\centering
\includegraphics[width=0.6\linewidth]{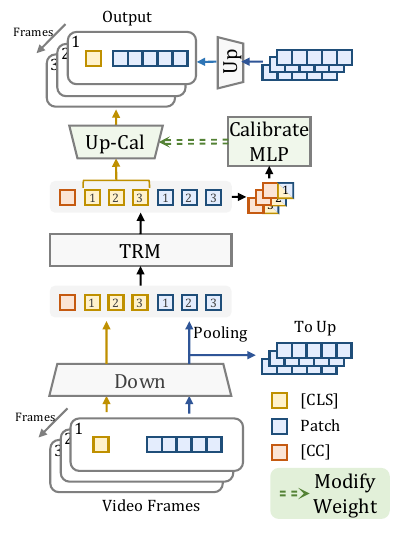}
\vspace{-3mm}\caption{Illustration of temporal adaptation in visual branch, including temporal modeling using lightweight transformer block (TRM) and temporal calibration to generate dynamic upsample weights for each frame.}
\label{fig:modules}
\vspace{-6mm}
\end{figure}
\subsection{Preliminary}
\label{subsec:preliminary}
In this sub-section, we briefly describe how VTR is performed by transferring from the pre-trained model CLIP \cite{CLIP}. We also introduce necessary notations used in the remainder of this paper. Like most state-of-the-art methods, CLIP is adopted due to its effectiveness and simple structure. Note our method can also be trivially extended to any CLIP-like backbone models that have the dual-encoder structure (for vision and text respectively). 

In VTR, the goal is to learn a relevance function ${\rm sim}(v, t)$ for calculating the similarity score between a video-text pair. In this way, given a video (text) query, we can obtain the rankings of all candidate texts (videos). Given a video-text pair $(v, t)$, we use the vision encoder $E_V$ and text encoder $E_T$ in CLIP for extracting features of $v$ and $t$ respectively. The depth of each encoder is denoted as $L$. We sample frames $(I_1, I_2, ... , I_{|v|})$ from $v$ to represent the video as a sequence of images. Then, each frame is patchified and prepended with a special [CLS] token. We pass them through $E_V$ and take the output of [CLS] token at the $L$-th layer as frame features $(E_{V}(I_1), E_{V}(I_2), ... , E_{V}(I_{|v|}))$. Finally, we obtain the global video features by aggregating frame features through mean-pooling:
$e_{v} = \frac{1}{|v|}\sum_{k=0}^{|v|} {E_{V}(I_k)}.$
Similarly, we pass the text $t$ to get its encoding $e_{t} = E_{T}(t)$. We calculate the cosine similarity between video feature and text feature
$
    {\rm sim}(v, t) = \frac{{e_{v}}^Te_{t}}{\|e_{v}\|||e_{t}\|}.
$

\subsection{Overview of MV-Adapter}
As illustrated in Fig.~\ref{fig:pipeline_overall}, MV-Adapter adds a new branch in the video/text encoder of CLIP respectively and bridges them through CMT module. Each branch is placed after the feed-forward network (FFN) in transformer block. Though there are substantial differences between their concrete forms, the basic structures of video/text branches are the same, following the bottleneck-like processing flow: \texttt{Downsample-Transformer-Upsample} 
 (Fig.~\ref{fig:pipeline_overall}). Both \texttt{Downsample} and \texttt{Upsample} are fully-connected layers, and the output is added to that of FFN. Formally, by denoting the output of FFN as $x \in \mathbb{R}^{N \times d}$ ($N$ and $d$ are the number and dimension of tokens respectively), the output of such an abstract structure is:
{\setlength\abovedisplayskip{0.5mm}
\setlength\belowdisplayskip{0.5mm}\begin{equation}
\label{eq:basic_text}
    A_{\rm basic}(x) = s\cdot \texttt{TRM}(xW_{\rm down})W_{\rm up},
\end{equation}}where $A_{\rm basic}()$ denotes the functions of the abstract structure we employ and $\texttt{TRM}()$ denotes the lightweight transformer. $W_{\rm down}\in\mathbb R^{d\times d^{\prime}}$ and $W_{\rm up}\in\mathbb R^{d^{\prime}\times d}$ are the weights of \texttt{Downsample} and \texttt{Upsample} respectively, $d^{\prime}$ is the feature dimension after downsampling and $s$ is scalar. For the text branch, we apply the process of Eq.~\eqref{eq:basic_text} directly where $N$ is the number of words tokens. For the video branch, we make significant modifications to the basic form to efficiently capture temporal cues. In addition, we propose a cross modality tying module with a shared parameter space to adjust weights for the \texttt{Downsample} in both modalities. We present more details of these components in the following sections.

\subsection{Temporal Adaptation}
\label{subsec:temp}
The goal of the video branch in MV-Adapter is to augment the image-text pre-trained vision encoder of CLIP with temporal modeling capability. To achieve this, as illustrated in \cref{fig:modules}, we encode temporal context into the spatial features of each frame and further enhance the model with dynamic temporal modeling.
 Before delving into details, we first expand the notation of the input of vision branch $x$ for description clarity. Considering all frames, $x=\{x^i\} _{i=1}^{|v|}$ where $x^i$ is the feature of $i$-th frame and $|v|$ is the number of frames. $x^i = [x_{\rm cls}^i, x^i_{\rm patch} ] \in \mathbb R^{(N_P + 1) \times d}$ where $x^i_{\rm patch} = [x^i_1, \cdots, x^i_{N_P}]$ and $N_P$ is the number of patches in each frame. Next, we detail how to enrich the image-level feature (for the [CLS] token)
 and patch-level features with temporal context. The pseudocode for the entire temporal adaption process is shown in Algorithm. \ref{algo:cls_cal}.
 \setlength{\textfloatsep}{1pt}
\begin{algorithm}[t]
\caption{Temporal Adaption}
\label{algo:cls_cal}
\vspace{-2mm}
\definecolor{codeblue}{rgb}{0.25,0.5,0.5}
\lstset{
  backgroundcolor=\color{white},
  basicstyle=\fontsize{6pt}{6pt}\ttfamily\selectfont,
  columns=fullflexible,
  breaklines=true,
  captionpos=b,
  commentstyle=\fontsize{6pt}{6pt}\color{codeblue},
  keywordstyle=\fontsize{6pt}{6pt},
  stringstyle=\fontsize{6pt}{6pt}\color{strorange},
  escapeinside={(*@}{@*)},
%  frame=tb,
}
\begin{lstlisting}[language=python]
# B: batchsize,         F: frame count,       L: sequence length
# E: feature dimension, E': middle dimension, P: patch number
# x: video features in visual encoder layer, shape is (B,F,L,E)
# cc: learnable parameter for video representation
# trm_block: lightweight transformer block
# cal_mlp: 2-layers MLP to generate calibrate weight
# down, up: down/up sample linear layer
# scale: the scaling factor when adding to the original features
def forward(x):
    # Downsample [CLS] and patch
    x_down = down(x)            # B,F,L,E'
    cls_seq = x_down[:,:,0]     # B,F,E'
    patch_down = x_down[:,:,1:] # B,F,P^2,E'
    
    # Temporal Sequence Modeling using transformer block 
    patch_seq = patch_down.mean(dim=2) # B,F,E'
    temporal_seq = concat([cc, cls_seq, patch_seq], dim=1)
    cc, cls_seq, patch_seq = trm_block(temporal_seq)
    
    # Temporal calibration weights ((*@\cref{eq:a_cal}@*)) 
    cc = cc[:, None].expand(-1, F, -1)  # B,F,E'
    cal_input = concat([cc, cls_seq + patch_seq], dim=-1) 
    alpha_cal_up = cal_mlp(cal_input)

    # Generate dynamic calibrated upsample weights
    # for each frame ((*@\cref{eq:w_up_cal}@*))
    up_w_cal = einsum('bfi,oi->bfio', alpha_cal_up, up.w)
    
    # Upsample process
    cls_up = einsum('bfio,bfi->bfo', up_w_cal, cls_seq)
    patch_up = up(patch_down)
    
    return x + scale * concat([cls_up, patch_up])
\end{lstlisting}
\vspace{-2mm}
\end{algorithm}

The [CLS] token's feature $x^i_{\rm [CLS]}$ is first passed through the $\texttt{Downsample}$ to reduce its dimension for computational efficiency. Then we concatenate all frames' [CLS] tokens and the average of patch tokens in the order of the frames in the video. And a learnable token, denoted as [CC] (``CC'' means the aggregated ``Class token'' of frame-wise ``Class tokens''), is appended to this sequence. This sequence is fed into the lightweight transformer $\texttt{TRM}$ as shown in Eq.~\eqref{eq:basic_text}. By utilizing the attention mechanism to capture the mutual dependencies among frame tokens, the transformer learns the temporal information across all frames. Consequently, each frame's [CLS] and averaged patch embeddings are enriched to be temporal-aware, and [CC] obtains global video representation, which will be used in the subsequent process to generate dynamic upsample weights for each frame.

To further enhance the temporal information on the [CLS] token, we design a temporal calibration module, which jointly use the video representation of ${x}_{\rm [CC]}$, each frame's [CLS] feature ${x}^i_{\rm [CLS]}$ and patch feature $\bar{x}^i_{\rm patch}$. Specifically, for $i$-th frame, we fuse above three features to obtain a calibration vector 
$\alpha^i=\texttt{concat}({x}_{\rm [CC]}, {x}^i_{\rm [CLS]}+\bar{x}^i_{\rm patch}) \in \mathbb R^{2d^\prime}$. Then, we generate the calibration weights $\alpha^i_{\rm cal} \in \mathbb{R}^{d^{\prime}}$ by feeding $\alpha^i$ into a two-layer calibration MLP: 
{\setlength\abovedisplayskip{0.5mm}
\setlength\belowdisplayskip{0.5mm}
\begin{equation}
    \alpha^i_{\rm cal} = \texttt{FC}_2(\texttt{ReLU}(\texttt{FC}_1(\alpha^i))),
\label{eq:a_cal}
\end{equation}}where $\texttt{FC}_1()$ and $\texttt{FC}_2()$ are fully connected layers whose weights dimensions are of  $\mathbb R^{2d^\prime\times (d^\prime/\sigma)}$ and $\mathbb R^{(d^\prime/\sigma)\times d^\prime}$ respectively. $\sigma$ is a shrinkage factor. Then we calibrate the weights of $\texttt{Upsample}$ using $\alpha^i_{\rm cal}$ as follows, 
{\setlength\abovedisplayskip{0.5mm}
\setlength\belowdisplayskip{0.5mm}
\begin{equation}
    (W_{\rm up\text{-}cal}^i)_c = \alpha_{\rm cal}^i\odot (W_{\rm up})_c,
\label{eq:w_up_cal}
\end{equation}}where $(W_{\rm up}^i)_c\in\mathbb{R}^{d}$ denotes the $c$-th column in $W_{\rm up}$. This calibration module enables the generation of individual dynamic weights for each frame, and $\texttt{Upsample}$ the [CLS] token of $i$-th frame using the calibrated weight $W^i_{up-cal}$.
\setlength{\textfloatsep}{1pt}
\begin{algorithm}[t]
\caption{Downsample using CMT}
\label{algo:cmt_cal}
\vspace{-2mm}
\definecolor{codeblue}{rgb}{0.25,0.5,0.5}
\lstset{
  backgroundcolor=\color{white},
  basicstyle=\fontsize{6pt}{6pt}\ttfamily\selectfont,
  columns=fullflexible,
  breaklines=true,
  captionpos=b,
  commentstyle=\fontsize{6pt}{6pt}\color{codeblue},
  keywordstyle=\fontsize{6pt}{6pt},
  stringstyle=\fontsize{6pt}{6pt}\color{strorange},
  escapeinside={(*@}{@*)},
%  frame=tb,
}
\begin{lstlisting}[language=python]
# E: feature dimension of current modality, E': middle dimension
# x: video features in visual encoder layer, shape is (*,E)
# down: downsample linear layer without modality interaction
# cmt_factor: cross modality factor in R^{d^m}
# cmt_proj: modality sepcific projector
def down(x):
    # project the shared factor, and generate modality-aware 
    # downsample weight down_w_cal ((*@\cref{eq:w_down_cal}@*)) 
    beta_cal = cmt_proj(cmt_factor)                
    down_w_cal = einsum('i,oi->oi', beta_cal, down.w)
    return x @ down_w_cal.T
\end{lstlisting}
\vspace{-2mm}
\end{algorithm}
For the patches in each frame $x^i_{\rm patch}$, the information density is not as concentrated as that of the [CLS] token, and the temporal information modeling can be obtained in the next layer through the interaction with the [CLS] token using the attention mechanism. Therefore, we directly pass them through $\texttt{Downsample}$ to reduce dimension, and then through $\texttt{Upsample}$ after activation to reduce computation.

 This process has two advantages. First, the weights change across different frames, allowing the model to capture the intricate variations of video dynamics. Second, since the video description $x_{\rm CC}$ , frame representations $x^i_{\rm CLS}$ and local feature $\bar{x}^i_{\rm patch}$ contain multi-grained context, features are enhanced with rich temporal contexts during the process. This strategy is considerably more appropriate for adapting to videos than using fixed weights in the $\texttt{Upsample}$ layer. Experimental results in \cref{tab:temporal} evidence the superiority of our method.

\subsection{Cross Modality Tying}
\label{subsec:cross}
Essentially, VTR brings the features of video and text closer in a joint embedding space. From this point of view, we design a cross modality tying (CMT) module for facilitating the alignment between modalities. Specifically, between the corresponding layers of visual and text encoders, we share a cross modality factor $f_C\in \mathbb R^{1\times d^m}$, where $d^m$ is the dimension of the shared factor. Subsequently, within each individual modality branch, we construct a modality specific projection matrix $M_S\in \mathbb R^{d^m\times d}$. During the $\texttt{Downsample}$ process, we utilize these parameters to adjust the weights of $\texttt{Downsample}$ across modalities. We employ the modality specific projection matrix to project the shared factors across modalities onto the feature dimension corresponding to the encoder of current modality, thereby obtaining the modality calibration vector $\beta^{\rm T/V}_{\rm cal}=f_{\rm C}\times M^{\rm T/V}_{\rm S}$, where $\rm{T/V}$ means Text or Visual branch. Then, we generated the modality-aware $\texttt{Downsample}$ weight using a method similar to that described in Eq. \ref{eq:w_down_cal} as follows:
{\setlength\abovedisplayskip{0.5mm}
\setlength\belowdisplayskip{0.5mm}\begin{equation}
    (W_{\rm down\text{-}cal}^{\rm T/V})_c = \beta^{\rm T/V}_{\rm cal}\odot (W_{\rm down})_c,
\label{eq:w_down_cal}
\end{equation}}
and $\texttt{Downsample}$ features with $W_{\rm down\text{-}cal}^{\rm T/V}$ rather than the original $\texttt{Downsample}$ weights. The pseudocode to describe the CMT process is shown in \cref{algo:cmt_cal}.

By sharing factor between modalities, CMT can implicitly reduce the distance in the embedding space, rather than forcibly aligning. Furthermore, it is a low-coupling method that allows the model to separately extract embeddings of videos and sentences, hence having much lower time complexity than other methods~\cite{kim2021vilt,c4c} (i.e., $O(\#\rm videos + \#\rm texts)$ vs. $O(\#\rm videos \times \#\rm texts)$) that require inputting video-text pairs together to obtain embeddings.
\subsection{Efficiency Analysis}
\noindent
\textbf{Deployment Efficiency} The MV-Adapter can significantly reduce the storage costs in deployment while maintaining performance, especially when serving a large amount of VTR tasks. This is due to its tunable parameter complexity of $O(2d\times d^{\prime}+k(d^{\prime})^2)$, where k is a constant that encompasses the parameter of both temporal calibration and TRM. MV-Adapter achieves this by storing the shared pre-trained model once and only a small number of tunable parameters for each task.  Put formally, suppose the size of pre-trained model is a unit, MV-Adapter reduces the space usage from $\#1\times \texttt{tasks}$ to $1 + \#\texttt{tasks}\times2.4\%$. Taken the five VTR benchmarks as examples, MV-Adapter can support all five tasks using \textbf{only 112\% }times the storage space of the pre-trained model, while the full fine-tuning and hunyuan\_tvr \cite{hy} takes {500\%} and about {600\%} respectively.

\noindent
\textbf{Training Efficiency} Besides, we also observe that MV-Adapter is more efficient and less computing resource demanding. It can considerably reduce around 40\% GPU memory costs compared to full fine-tuning, as the majority parts of model are frozen. We argue that this appealing property of MV-Adapter makes various training optimizations feasible, such as using a larger batch size to improve contrastive learning \cite{he2020momentum,chen2020improved,wu2018unsupervised}.

%% file: sec/4_experiment.tex
\section{Experiment Setup}
\subsection{Datasets and Evaluation Metric}
We evaluate our module on five commonly used public video-text retrieval datasets: MSR-VTT, MSVD, LSMDC, DiDemo and ActivityNet.
\noindent\\
{\bf MSR-VTT \cite{msrvtt} }is widely used in previous literature, consisting of 10,000 videos with 20 captions each. The training set we use has 9k video-text pairs \cite{msrvtt9k}, and the test set has 1k video-text pairs \cite{msrvtt1k}. \\
{\bf MSVD \cite{msvd}} contains 1970 videos with approximately 40 captions each. Train, validate and test sets have 1200, 100, and 670 videos respectively.\\
{\bf LSMDC \cite{lsmdc}} contains 118081 video-text pairs, and all the videos are extracted from 202 movies. The movie sources for the training and test videos are independent.\\
{\bf DiDemo \cite{didemo}} contains 10000 videos, and each has four sentences concatenated as the whole caption following \cite{c4c}. \\
{\bf ActivityNet \cite{anet}} has 20000 Youtube videos. Similarly as in~\cite{didemo}, we combine all sentences as the final text.

We use the standard retrieval metrics: recall at rank K (R@K, higher is better) following previous practices. To eliminate randomness, we report both the mean and std of results on three seeds (0, 42, 123) for each setup. We also report the sum of R@1/5/10 (higher is better).
\subsection{Implementation Detail}
We adopt CLIP (ViT-B/16) \cite{CLIP} as the backbone of our model by default and ablate using CLIP (ViT-L/14) to validate the scalability of MV-Adapter. The parameters of adapter are optimized by Adam, while the learning rate is 5e-6. For MSR-VTT, MSVD, and LSMDC datasets, the max lengths of frames and words in captions are 12 and 32, and we train the module for 5 epochs. For ActivityNet and DiDemo datasets, we set the lengths of video and caption to 32 and 64 respectively as they are longer and use 15 epochs. The batch size is 128 by default and 64 for Didemo and ActivityNet due to GPU memory limit. Following \cite{c4c}, we extract 1 frame per second from the videos and select the frames uniformly when the number of frames is larger than the max length. The middle dimension of MV-Adapter is 64, and the shrinkage factor $\sigma$ and $s$ are set as 4, 0.1.

%% file: sec/5_result.tex
\section{Results And Analysis}
\begin{table*}
  \centering
  \scriptsize
  \begin{tabular}{l|c|ccc|c|ccc|c@{}}
    \toprule
    \multirow{2}*{Method} & Tunable  &\multicolumn{3}{c|}{Text-to-Video} & & \multicolumn{3}{c|}{Video-to-Text} & \\
    & Params(\%) & R@1 &  R@5 &  R@10 & Sum & R@1 &  R@5 &  R@10 & Sum \\
    \midrule
    Full Fine-tuning & 100 & 45.0 & \underline{73.0} & \underline{82.2} & \underline{200.2} & {45.3} & \underline{73.4} & \underline{83.7} & \underline{202.4} \\
    \midrule
    $\mathrm{Adapt}_{par}$ \cite{adaptmlp} & 2.78 & 45.5 $\pm$ 0.2 & 72.3 $\pm$ 0.4 & 81.5 $\pm$ 0.5 & 199.3 & \underline{45.7} $\pm$ 0.2 & 72.9 $\pm$ 0.2 & 82.7 $\pm$ 0.3 & 201.3 \\
    
    $\mathrm{Adapt}_{seq}$ \cite{adaptmlp}  & 2.78 & \underline{45.7} $\pm$ 0.1 & 72.4 $\pm$ 0.4 & 81.0 $\pm$ 0.2 & 199.1 & 45.2 $\pm$ 0.2 & 72.8 $\pm$ 0.9 & 81.7 $\pm$ 0.6 & 199.7 \\
    
    Convpass \cite{convpass} & 2.80 & 42.5 $\pm$ 0.5 & 69.8 $\pm$ 0.2 & 80.4 $\pm$ 0.1 & 192.6 & 44.7 $\pm$ 1.0 & 72.5 $\pm$ 0.1 & 82.1 $\pm$ 0.2 & 199.4 \\
    
    ST \cite{st} & 5.93 & 43.6 $\pm$ 0.5 & 70.9 $\pm$ 0.5 & 81.4 $\pm$ 0.7 & 196.0 & 45.5 $\pm$ 0.8 & 72.7 $\pm$ 0.2 & 82.7 $\pm$ 0.6 & 201.0 \\
    CM \cite{cm_thu}$^*$ & \underline{2.76} & 42.1 $\pm$ 0.4 & 69.5 $\pm$ 0.3 & 79.7 $\pm$ 0.2 & 191.3 & 42.0 $\pm$ 0.6 & 69.9 $\pm$ 0.3 & 80.7 $\pm$ 0.5 & 192.6\\

    \midrule
    CLIP4Clip \cite{c4c} & 8.45 & 42.1 & 68.3 & 78.8 & 189.2 & 40.2 & 68.1 & 79.1 & 187.4\\
    Hunyuan \cite{hy}$^*$ & 11.97 & 43.8 $\pm$ 0.3 & 70.9 $\pm$ 0.7 & 81.1 $\pm$ 0.6 & 195.8 & 41.2 $\pm$ 1.0 & 70.5 $\pm$ 0.6 & 80.6 $\pm$ 0.6 & 192.3\\

    \midrule
    MV-Adapter & \textbf{2.39} & \textbf{46.2} $\pm$ 0.5 & \textbf{73.2} $\pm$ 0.3 & \textbf{82.7} $\pm$ 0.3 & \textbf{202.1} & \textbf{47.2} $\pm$ 0.4 & \textbf{74.8} $\pm$ 0.3 & \textbf{83.9} $\pm$ 0.5 & \textbf{205.9}\\
    \bottomrule
  \end{tabular}
  \vspace{-2.5mm}
  \caption{Comparison results on MSR-VTT \cite{msrvtt} using CLIP (ViT-B/16). \textbf{bold} and \underline{underline} indicates the top two results. $^*$ denotes our reproduced results.}
  \label{tab:res_msrvtt}
  \vspace{-2mm}
\end{table*}
%msvd
\begin{table*}
  \centering
  \scriptsize
  \begin{tabular}{l|c|ccc|c|ccc|c@{}}
    \toprule
    \multirow{2}*{Method} & Tunable  &\multicolumn{3}{c|}{Text-to-Video} & & \multicolumn{3}{c|}{Video-to-Text} & \\
    & Params(\%) & R@1 &  R@5 &  R@10 & Sum & R@1 &  R@5 &  R@10 & Sum \\
    \midrule
    Full Fine-tuning & 100 & \textbf{49.7} & \textbf{79.2} & \textbf{87.3} & \textbf{216.2} & \underline{71.2} & \underline{92.5} & \underline{95.5} & \underline{259.3} \\
    \midrule
    $\mathrm{Adapt}_{par}$ \cite{adaptmlp} & 2.78 & 48.3 $\pm$ 0.1 & 77.7 $\pm$ 0.0 & 86.8 $\pm$ 0.0 & 212.8 & 68.9 $\pm$ 0.3 & 90.9 $\pm$ 0.6 & 94.8 $\pm$ 0.2 & 254.6 \\
    
    $\mathrm{Adapt}_{seq}$ \cite{adaptmlp} & 2.78 & 45.1 $\pm$ 0.2 & 76.2 $\pm$ 0.2 & 85.3 $\pm$ 0.1 & 206.7 & 62.6 $\pm$ 0.6 & 88.8 $\pm$ 0.2 & 93.4 $\pm$ 0.4 & 244.8 \\
    
    Convpass \cite{convpass} & 2.80 & 46.1 $\pm$ 0.3 & 76.0 $\pm$ 0.3 & 85.5 $\pm$ 0.1 & 207.6 & 70.0 $\pm$ 1.2 & 91.9 $\pm$ 0.6 & \underline{95.5} $\pm$ 0.6 & 257.4 \\
    
    ST \cite{st} & 5.93 & 45.9 $\pm$ 0.2 & 75.9 $\pm$ 0.2 & 85.2 $\pm$ 0.2 & 207.0 & 69.9 $\pm$ 1.6 & 91.7 $\pm$ 1.0 & 95.1 $\pm$ 0.4 & 256.7 \\
    CM \cite{cm_thu}$^*$ & \underline{2.76} & 44.7 $\pm$ 0.8 & 75.8 $\pm$ 0.4 & 85.0 $\pm$ 0.3 & 205.5 & 61.5 $\pm$ 0.9 & 88.2 $\pm$ 0.5 & 93.2 $\pm$ 0.6 & 242.9\\
    \midrule
    CLIP4Clip \cite{c4c} & 8.45 & 45.0 & 74.2 & 83.4 & 202.5 & 62.2 & 88.7 & 94.0 & 244.9\\
    Hunyuan \cite{hy}$^*$ & 11.97 & 43.0 $\pm$ 0.4 & 73.7 $\pm$ 0.2 & 83.6 $\pm$ 0.2 & 200.3 & 53.7 $\pm$ 2.1 & 83.9 $\pm$ 2.2 & 90.2 $\pm$ 1.5 & 227.8\\
    
    \midrule
    MV-Adapter & \textbf{2.39} & \underline{49.4} $\pm$ 0.2 & \underline{78.3} $\pm$ 0.1 & \underline{87.0} $\pm$ 0.1 & \underline{214.8} & \textbf{71.8} $\pm$ 0.5 & \textbf{93.0} $\pm$ 0.4 & \textbf{96.4} $\pm$ 0.1 & \textbf{261.2}\\
    \bottomrule
  \end{tabular}
  \vspace{-2.5mm}
  \caption{Comparison results on MSVD \cite{msvd} using CLIP (ViT-B/16). \textbf{bold} and \underline{underline} indicates the top two results. $^*$ denotes our reproduced results.}
  \label{tab:res_msvd}
  \vspace{-6mm}
\end{table*}
\subsection{Baselines}
\label{sec:baselines}
As a parameter-efficient method, we first compare with full fine-tuning. To further verify the effectiveness of our method, we build strong baselines by adapting popular vision adapters to the VTR task. The details are as follows.
\noindent\\
{\bf Full fine-tuning.} Update all parameters for each task, which refers to the ``MeanP'' Method in \cite{c4c}.\\
{\bf AdaptMLP~\cite{adaptmlp}.} This module is a bottleneck block containing two fully connected layers $W_{\rm down}\in\mathbb R^{d\times d^{\prime}}$ and $W_{\rm up}\in\mathbb R^{d^{\prime}\times d}$, where $d$ is the feature dimension and $d^{\prime} \ll d$. AdaptMLP has two forms according to its location in the transformer block. In \textit{parallel} form, it is placed after \texttt{FFN} and takes the form of 
{\setlength\abovedisplayskip{0.1cm}
\setlength\belowdisplayskip{0.1cm}\begin{equation*}
A(x)=x + s\cdot\texttt{ReLU}(x W_{\rm down}) W_{\rm up},
\end{equation*}}where $x$ is the input of \texttt{FFN} as in Eq.~\eqref{eq:basic_text}. In \textit{sequential} form, it has the form of {\setlength\abovedisplayskip{0.1cm}
\setlength\belowdisplayskip{0.1cm}\begin{equation*}
A(x)=\texttt{FFN}(x) + s\cdot\texttt{ReLU}(\texttt{FFN}(x) W_{\rm down}) W_{\rm up}.
\end{equation*}Due to its simplicity, AdaptMLP can be conveniently applied to transformer structures. Therefore, we use sequential AdaptMLP (for its better performance) in the text encoder to tailor the following vision adapters to VTR. We use Adapt$_{par/seq}$ to represent these forms in result tables.\\
\noindent
{\bf Convpass~\cite{convpass}. } It uses convolution for adaptation  in vision domains. Specifically, the patch features are separated from the [CLS] features and reshaped to the 2D matrix according to their spatial locations. Then, both the [CLS] and patches features are fed into Convpass respectively
{\setlength\abovedisplayskip{0.1cm}
\setlength\belowdisplayskip{0.1cm}\begin{equation*}
A(x)=\texttt{FFN}/\texttt{MHSA}(x) + s\cdot\texttt{Conv}_{3\times 3}(x W_{\rm down}) W_{\rm up}.
\end{equation*}}{\bf ST-Adapter~\cite{st}. } This module is for adapting image model to video tasks, employed at the beginning of each transformer block. It uses a depth-wise 3D-convolution to capture Spatio-Temporal information as follows {\setlength\abovedisplayskip{0.1cm}
\setlength\belowdisplayskip{0.1cm}\begin{equation*}
A(x)=x + \texttt{DWConv}(x W_{\rm down}) W_{\rm up}.
\end{equation*}}{\bf CM-Adapter~\cite{cm_thu}. } Based on AdaptMLP~\cite{adaptmlp}, it simply appends a shared learnable weight between encoders. The module is inserted to each layer of CLIP.
{\setlength\abovedisplayskip{0.1cm}
\setlength\belowdisplayskip{0.1cm}\begin{equation*}
    A(x)=x+\sigma(xW_{\rm down})\texttt{Concat}[W_{\rm up}, W_{\rm up}^{\rm cm}],
\end{equation*}}where $\sigma(\cdot)$ is activation function.

For the above methods, we add and train AdaptMLP in their text encoders to ensure fair comparison. AdaptMLP is selected for its wide use and effectiveness across modalities. Furthermore, we adapt the state-of-the-art VTR methods using full fine-tuning:  Hunyuan~\cite{hy} and CLIP4Clip (+SeqTransformer)~\cite{c4c} to PE-VTR by freezing the update of CLIP during training. In this way, they can be regarded as adapters to the output of the last layer.
%lsmdc
\begin{table*}[!ht]
  \centering
  \scriptsize
  \begin{tabular}{l|c|ccc|c|ccc|c@{}}
    \toprule
    \multirow{2}*{Method} & Tunable  &\multicolumn{3}{c|}{Text-to-Video} & & \multicolumn{3}{c|}{Video-to-Text} & \\
    & Params(\%) & R@1 &  R@5 &  R@10 & Sum & R@1 &  R@5 &  R@10 & Sum \\
    \midrule
    Full Fine-tuning & 100 & 20.2 & \underline{41.5} & \underline{51.2} & \underline{112.9} & 21.8 & \underline{40.5} & \underline{50.9} & \underline{113.2} \\
    \midrule
    $\mathrm{Adapt}_{par}$ \cite{adaptmlp} & 2.78 & 19.8 $\pm$ 0.4 & 39.0 $\pm$ 0.3 & 48.8 $\pm$ 0.5 & 107.6 & 21.0 $\pm$ 0.1 & 38.0 $\pm$ 0.7 & 46.6 $\pm$ 0.9 & 105.5 \\
    
    $\mathrm{Adapt}_{seq}$ \cite{adaptmlp} & 2.78 & 20.7 $\pm$ 0.5 & 40.2 $\pm$ 0.2 & 50.2 $\pm$ 0.4 & 111.1 & 20.9 $\pm$ 0.1 & 38.8 $\pm$ 0.5 & 49.1 $\pm$ 0.4 & 108.8 \\
    
    Convpass \cite{convpass} & 2.80 & 19.7 $\pm$ 0.1 & 37.8 $\pm$ 0.4 & 46.3 $\pm$ 0.5 & 103.8 & 20.7 $\pm$ 0.5 & 38.9 $\pm$ 0.3 & 47.7 $\pm$ 0.7 & 107.2 \\
    
    ST \cite{st} & 5.93 & \underline{20.9} $\pm$ 0.6 & 39.8 $\pm$ 0.4 & 49.1 $\pm$ 0.9 & 109.8 & \underline{21.9} $\pm$ 0.2 & 40.3 $\pm$ 0.3 & 49.5 $\pm$ 0.6 & 111.8 \\
    CM \cite{cm_thu}$^*$ & \underline{2.76} & 18.7 $\pm$ 0.5 & 38.7 $\pm$ 0.3 & 48.3 $\pm$ 0.1 & 105.7 & 20.9 $\pm$ 0.3 & 37.8 $\pm$ 1.0 & 47.7 $\pm$ 0.5 & 106.3\\

    \midrule
    CLIP4Clip \cite{c4c} & 8.45 & 20.1 & 37.4 & 46.0 & 103.5 & 18.1 & 34.6 & 43.9 & 96.6 \\
    Hunyuan \cite{hy}$^*$ & 11.97 & 20.6 $\pm$ 0.3 & 37.6 $\pm$ 0.5 & 46.7 $\pm$ 0.6 & 104.9 & 18.5 $\pm$ 0.8 & 36.6 $\pm$ 0.5 & 44.5 $\pm$ 0.8 & 99.6 \\
    
    \midrule
    MV-Adapter & \textbf{2.42} & \textbf{23.2} $\pm$ 0.7 & \textbf{43.9} $\pm$ 0.5 & \textbf{53.2} $\pm$ 0.6 & \textbf{120.3} & \textbf{24.0} $\pm$ 0.5 & \textbf{42.8} $\pm$ 0.4 & \textbf{52.1} $\pm$ 0.2 & \textbf{118.8}\\
    \bottomrule
  \end{tabular}
  \vspace{-2.5mm}
  \caption{Comparison results on LSMDC~\cite{lsmdc} using CLIP (ViT-B/16). \textbf{bold} and \underline{underline} indicates the top two results. $^*$ denotes our reproduced results.}
  \label{tab:res_lsmdc}
  \vspace{-2mm}
\end{table*}
% didemo
\begin{table*}[!ht]
  \centering
  \scriptsize
  \begin{tabular}{l|c|ccc|c|ccc|c@{}}
    \toprule
    \multirow{2}*{Method} & Tunable  &\multicolumn{3}{c|}{Text-to-Video} & & \multicolumn{3}{c|}{Video-to-Text} & \\
    & Params(\%) & R@1 &  R@5 &  R@10 & Sum & R@1 &  R@5 &  R@10 & Sum \\
    \midrule
    Full Fine-tuning & 100 & \textbf{44.7} & \textbf{73.6} & \textbf{81.2} & \textbf{199.6} & \textbf{45.0} & \textbf{73.1} & \underline{80.8} & \textbf{198.9} \\
    \midrule
    $\mathrm{Adapt}_{par}$ \cite{adaptmlp} & 2.78 & 42.6 $\pm$ 0.2 & \underline{72.5} $\pm$ 0.4 & \underline{80.8} $\pm$ 0.5 & 195.9 & \underline{43.0} $\pm$ 0.0 & 70.6 $\pm$ 0.4 & 80.0 $\pm$ 0.7 & 193.5 \\
    
    $\mathrm{Adapt}_{seq}$ \cite{adaptmlp} & 2.78 & 42.4 $\pm$ 0.8 & 70.6 $\pm$ 0.4 & 80.3 $\pm$ 0.1 & 193.3 & 42.2 $\pm$ 0.6 & 69.9 $\pm$ 0.3 & 79.7 $\pm$ 0.2 & 191.8 \\
    
    Convpass \cite{convpass} & 2.80 & 40.7 $\pm$ 1.1 & 69.6 $\pm$ 0.5 & 78.2 $\pm$ 0.7 & 188.6 & 40.9 $\pm$ 0.3 & 70.0 $\pm$ 0.8 & 79.2 $\pm$ 1.0 & 190.0 \\
    
    ST \cite{st} & 5.93 & 40.4 $\pm$ 0.1 & 69.4 $\pm$ 0.6 & 79.2 $\pm$ 0.3 & 189.0 & 41.2 $\pm$ 1.1 & 70.1 $\pm$ 0.4 & 80.1 $\pm$ 0.5 & 191.4 \\
    CM \cite{cm_thu}$^*$ & \underline{2.77} & 42.9 $\pm$ 0.5 & 70.6 $\pm$ 0.3 & 80.2 $\pm$ 0.1 & 193.7 & 42.5 $\pm$ 1.0 & 70.8 $\pm$ 0.4 & 80.5 $\pm$ 0.2 & 193.8 \\

    \midrule
    CLIP4Clip \cite{c4c} & 8.45 & 36.2 & 61.8 & 72.7 & 170.7 & 34.4 & 62.4 & 72.9 & 169.7\\
    Hunyuan \cite{hy}$^*$ & 11.97 & 37.0 $\pm$ 0.6 & 64.2 $\pm$ 1.2 & 74.3 $\pm$ 1.0 & 175.5 & 34.6 $\pm$ 1.0 & 63.0 $\pm$ 1.1 & 73.6 $\pm$ 0.3 & 171.2 \\

    \midrule
    MV-Adapter & \textbf{2.42} & \underline{44.3} $\pm$ 0.4 & 72.1 $\pm$ 0.6 & 80.5 $\pm$ 0.2 & \underline{196.8} & 42.7 $\pm$ 1.0 & \underline{73.0} $\pm$ 0.7 & \textbf{81.9} $\pm$ 0.5 & \underline{197.6} \\
    \bottomrule
  \end{tabular}
  \vspace{-2mm}
  \caption{Comparison results on DiDemo~\cite{didemo} using CLIP (ViT-B/16). \textbf{bold} and \underline{underline} indicates the top two results. $^*$ denotes our reproduced results.}
  \label{tab:res_didemo}
  \vspace{-2mm}
\end{table*}
%anet
\begin{table*}[!ht]
  \centering
  \scriptsize
  \begin{tabular}{l|c|ccc|c|ccc|c@{}}
    \toprule
    \multirow{2}*{Method} & Tunable  &\multicolumn{3}{c|}{Text-to-Video} & & \multicolumn{3}{c|}{Video-to-Text} & \\
    & Params(\%) & R@1 &  R@5 &  R@10 & Sum & R@1 &  R@5 &  R@10 & Sum \\
    \midrule
    Full Fine-tuning & 100 & \underline{42.9} & 73.2 & 85.4 & {201.5} & \textbf{43.8} & \textbf{75.0} & \textbf{86.6} & \textbf{205.3} \\
    \midrule
    $\mathrm{Adapt}_{par}$ \cite{adaptmlp} & 2.78 & 41.7 $\pm$ 0.0 & 72.7 $\pm$ 0.2 & 84.5 $\pm$ 0.1 & 198.8 & 43.3 $\pm$ 0.0 & 73.1 $\pm$ 0.1 & 85.8 $\pm$ 0.1 & 202.2 \\
    $\mathrm{Adapt}_{seq}$ \cite{adaptmlp} & 2.78 & 41.2 $\pm$ 0.1 & 72.2 $\pm$ 0.1 & 84.1 $\pm$ 0.0 & 197.5 & 42.5 $\pm$ 0.2 & 73.0 $\pm$ 0.2 & 85.2 $\pm$ 0.1 & 200.7 \\
    % 40.9 $\pm$ 0.1 & 72.1 $\pm$ 0.2 & 84.4 $\pm$ 0.3 & 197.4 & 41.0 $\pm$ 0.1 & 71.2 $\pm$ 0.2 & 84.2 $\pm$ 0.2 & 196.4 \\
    
    Convpass \cite{convpass} & 2.80 & 38.6 $\pm$ 0.3 & 69.9 $\pm$ 0.5 & 82.7 $\pm$ 0.3 & 191.2 & 40.5 $\pm$ 0.1 & 71.9 $\pm$ 0.3 & 84.1 $\pm$ 0.1 & 196.6 \\
    
    ST \cite{st} & 5.93 & 38.7 $\pm$ 0.3 & 69.8 $\pm$ 0.3 & 83.1 $\pm$ 0.3 & 191.6 & 40.7 $\pm$ 0.3 & 71.8 $\pm$ 0.3 & 84.2 $\pm$ 0.4 & 196.7 \\
    CM \cite{cm_thu}$^*$ & \textbf{2.25} & \textbf{43.1} $\pm$ 0.2 & \underline{74.2} $\pm$ 0.1 & \underline{85.5} $\pm$ 0.1 & \underline{202.8} & 43.5 $\pm$ 0.5 & 74.4 $\pm$ 0.1 & 86.2 $\pm$ 0.2 & 204.0\\

    \midrule
    CLIP4Clip \cite{c4c} & 8.45 & 36.9 & 68.5 & 8.01 & 186.4 & 35.1 & 68.4 & 82.0 & 185.5 \\
    Hunyuan \cite{hy}$^*$ & 11.97 & 37.8 $\pm$ 0.3 & 70.4 $\pm$ 0.4 & 83.5 $\pm$ 0.2 & 191.7 & 35.1 $\pm$ 0.2 & 69.1 $\pm$ 0.9 & 83.0 $\pm$ 0.5 & 187.2 \\
    
    \midrule
    MV-Adapter & \underline{2.40} & \underline{42.9} $\pm$ 0.0 & \textbf{74.5} $\pm$ 0.1 & \textbf{85.7} $\pm$ 0.1 & \textbf{203.1} & \underline{43.6} $\pm$ 0.1 & \textbf{75.0} $\pm$ 0.3 & \underline{86.5} $\pm$ 0.1 & \underline{205.2} \\
    \bottomrule
  \end{tabular}
  \vspace{-2.5mm}
  \caption{Comparison results on ActivityNet~\cite{anet} using CLIP (ViT-B/16). \textbf{bold} and \underline{underline} indicates the top two results. $^*$ denotes our reproduced results.}
  \label{tab:res_anet}
  \vspace{-4mm}
\end{table*}
\subsection{Main Result}
\label{subsec:result}
In this section, we present in the detail the comparison of MV-Adapter with other methods
on MSR-VTT, MSVD, LSMDC, Didemo, and ActivityNet. The results of experiments are shown in \cref{tab:res_msrvtt,tab:res_msvd,tab:res_lsmdc,tab:res_didemo,tab:res_anet} (note that the results of full fine-tuning and CLIP4Clip with SeqTransf are constant, as the initialization is fixed). We first compare our method with baseline method which uses standard full fine-tuning. Overall, we find our method performs on par or even better (in most cases) than the baseline. For example, on the T2V task, MV-Adapter surpasses the full fine-tuning by 1.9, 7.4 and 1.6 on the sum of R@1/5/10 on MSR-VTT, LSMDC and ActivityNet respectively. Similarly, on V2T, MV-adapter outperforms the full fine-tuning by 3.5, 1.9, and 5.6 on MSR-VTT, MSVD, and LSMDC. 

Afterwards, we adapt previous PETL methods including AdaptFormer \cite{adaptmlp}, Convpass \cite{convpass} and ST-adapter \cite{st} to VTR as introduced in Sec.~\ref{sec:baselines}. Also, we compare with state-of-the-art methods in the full fine-tuning setting by freezing the CLIP backbone: CLIP4Clip (+seqrTransformer) \cite{c4c} and Hunyuan \cite{hy}. As shown in \cref{fig:vis}, MV-Adapter consistently outperforms other methods with the smallest parameter overhead. Specifically, compared with other methods, MV-Adapter shows signficant improvements. Measured by the R@Sum of T2V/V2T, MV-Adapter outperforms the second best method by 6.1/4.9, 2.0/6.6, 10.5/7.0, 0.9/4.1 and 0.3/1.2 on MSR-VTT, MSVD, LSMDC, Didemo and ActivityNet, respectively. MV-Adapter achieves this with about 60\% of the GPU memory usage as the tunable parameters are small.

\subsection{Ablations}
\label{sec:ablations}
\begin{figure*}[h]
\centering
\includegraphics[width=\linewidth]{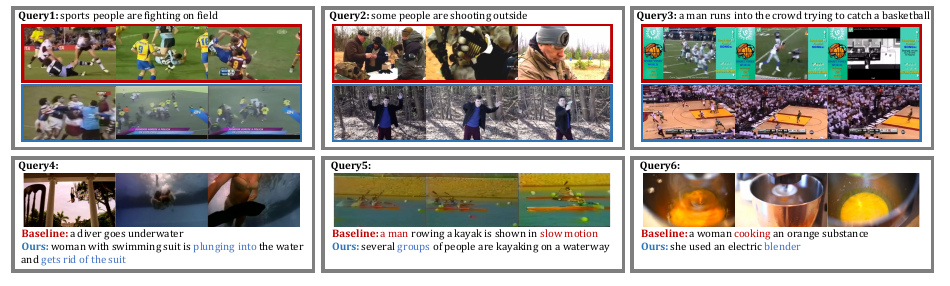}
\vspace{-9mm}
\caption{Visualizations of text-to-video (top row) and video-to-text (bottom row) results from MV-Adapter and ST-Adapter~\cite{st} using the same query from MSR-VTT. 
In each example, the retrieval results of baseline and MV-Adapter are shown in \textcolor[RGB]{192,0,0}{red} and \textcolor[RGB]{46,117,182}{blue} respectively.}
\vspace{-2mm}
\label{fig:vis}
\end{figure*}

\begin{table*}
\centering

\footnotesize
    % temporal information
    \begin{subtable}[h]{0.3\textwidth}
      \centering
      \begin{tabular}{l|cc}
        Settings & T2V & V2T  \\
        \midrule
        Full FT & 45.0/200.2 & 45.3/202.4\\
        Down\&Up & 42.9/192.0 & 43.4/195.3\\
        w/ TRM & 45.0/197.5 & 46.0/202.9  \\
        w/ calibration & \textbf{45.9/200.9} & \textbf{46.6/204.4} \\
      \end{tabular}
      \label{tab:temporal}
    \end{subtable}
        \hfill
    \centering
    \begin{subtable}[h]{0.35\textwidth}
      \centering
      \begin{tabular}{ccc|cc}
        Visual & Text & CMT & T2V & V2T  \\
        \midrule

         & \usym{2713} & & 44.0/193.1 & 45.2/199.1 \\
        \usym{2713} & & & 44.2/191.7 & 42.0/192.3  \\
        \usym{2713} & \usym{2713} & & 45.9/200.9 & 46.6/204.4 \\
        \usym{2713} & \usym{2713} & \usym{2713} & \textbf{46.2/202.1} & \textbf{47.2/205.9}\\
      \end{tabular}
      \label{tab:modality}
    \end{subtable}
    %2022
    \hfill
    \centering
    \begin{subtable}[c]{0.3\textwidth}
    \centering
      \begin{tabular}{l|cc}
        Method & T2V & V2T \\
        \midrule
        Full FT & 49.2/208.0 & 49.2/211.7 \\
        Ours & \textbf{49.3/209.3} & \textbf{49.6/212.4} \\

      \end{tabular}
    \end{subtable}
    \vspace{-3mm}
    \caption{Ablations. We report R@1 and the R@sum on MSR-VTT, where ``Full FT'' means Full Fine-tuning Method. Each reported result represents the average derived from experiments conducted on three different seeds (0, 42, 123). Due to space limit, please refer to more ablations in the Appendix. (a) Left. Performance improvements brought by adding temporal modeling and then adding temporal calibration. (b) Middle. Analysis of using video/text branches and CMT. (c) Right. Scalability analysis based on CLIP (ViT-L/14).}
    \label{tab:ablation_}
    \vspace{-2.8mm}
\end{table*}

We conducted extensive ablative experiments to validate the effectiveness of the design choices of MV-Adapter. In this section, we measure the model on MSR-VTT\cite{msrvtt}. For simplicity, we use the R@1 and the sum of R@1/5/10 averaged on three seeds as the metric. \\
\noindent
{\bf Temporal Adaptation.} 
In this experiment, we investigate the effects of the proposed temporal adaptation module. Results are listed on \cref{tab:temporal}. First, we construct a baseline model that simply applies the bottleneck structure. This model achieves only 42.9/43.4 T2V/V2T R@1 performance which is much worse than the performance of full fine-tuning (45.7/45.3). Then we add the temporal modeling using transformer block as introduced in Sec.~\ref{subsec:temp}, the performance significantly improves to 45.0/46.0. This verifies the importance of learning temporal contexts across frames. Furthermore, with the temporal calibration, the model is boosted to 45.9/46.6, outperforming the fully fine-tuned model. Above results strongly demonstrate the effectiveness of temporal adaptation.\vspace{2.5mm}\\
\noindent
{\bf Multimodal Branches and CMT.} Since VTR inherently involves learning from two modalities, it is natural to assume that multimodal adaptation is more suitable for the task. To corroborate this, we conduct experiments using different combinations of video/text branches. As shown in the Tab.~\ref{tab:ablation_} (b), applying only text branch or video branch all decreases the performances (-7.8/5.3 and -9.2/12.1 on the T2V/V2T R@Sum respectively), and CMT further improves the performance of the model by 1.2/1.5  on the T2V/V2T R@Sum. These results evidently support the utilization of both branches and CMT in MV-Adapter.\\
\noindent
{\bf Model Scalability.} In order to validate the scalability of MV-Adapter, we use the CLIP V-L/14 as the backbone to perform the ablation study. Since CLIP V-L/14 has a higher feature dimension in the encoder than its base version, we adjust the middle dimension of MV-Adapter to 128 to accommodate the larger backbone. We first obtain fully fine-tuned results as the baseline, then experiment with MV-Adapter. The experimental results in Tab. \ref{tab:ablation_} (c) are consistent with those of CLIP V-B/16, where MV-Adapter achieves better results with small parameter overhead.\\
\noindent{\bf More ablations.} Due to space limit, please refer to the Appendix for ablations on other datasets and ablations of the validity and locations of CMT.
\noindent
\subsection{Qualitative Results}
In order to better understand how MV-Adapter performs compared with other methods, we illustrate some qualitative results on V2T and T2V tasks in Fig.~\ref{fig:vis}. We use ST-Adapter~\cite{st} for comparison since its performance on MSR-VTT is only worse than ours. As can be observed in Fig.~\ref{fig:vis}, MV-Adapter is capable of modeling rich spatio-temporal information in videos including dynamic movements and relationships among objects, thus building more accurate correspondence between video and text. 
In contrast, the baseline method lacks in temporal modeling capabilities, thus failing to comprehend complex video scenes. For example, in \texttt{Query1} (T2V), the video returned by MV-Adapter accurately understand the complete action of fighting, while the baseline method erroneously returns clips of competitive sports confrontation. Similarly, when using video \texttt{Query4} (V2T) for retrieval, MV-Adapter can correctly understand the action and the entire temporal process. In comparison, the baseline method only attends to the main scene of the video, which is swimming underwater, without understanding the correlation before and after.

We summarize two failure modes from case studies. (1) Since we use a fixed number of frames per video in training, it may fail to capture fine movements in long videos as the differences between frames aggregate. Employing a sliding window of videos may be helpful. (2) Results may be incorrect when the caption is related to the audio aspect. Performing ASR or augmenting the input with audio features may solve this issue. Examples will be shown in Appendix.

%% file: sec/6_conclustion.tex
\section{Conclusion}
{\linespread{1}\selectfont In this paper, we propose the task of parameter-efficient VTR (PE-VTR) to save storage costs of models. Previous methods fail to address PE-VTR, leading to inferior performance and/or large parameter overheads. To tackle PE-VTR, we introduce MV-Adapter with two novel components: temporal adaptation module and cross modality tying. MV-Adapter achieves comparable results with full fine-tuning and outperforms competing methods. In the future, we plan to extend MV-Adapter to more multimodal tasks such as video question answering and captioning.\par}

%% file: sec/X_suppl.tex
\clearpage
\setcounter{page}{1}

\section{More Ablations}
\label{sec:ablations}
% \begin{table*}[hb]
% \centering
% \end{table*}
\noindent {\bf Cross Modality Tying.} 
We conduct extensive experiments to validate the design and settings of CMT.
\begin{itemize}
    \item \textbf{Design.} As shown in \cref{eq:w_down_cal}, CMT calibrates the weights of \texttt{Downsample} by element-wise multiplication of $\beta_{\text{cal}} \in \mathbb{R}^d$ with each column of $W_{\rm down}$. Here, the dimension of $\beta_{\text{cal}}$ is equal to the in-channels of $W_{\rm down}$, which we refer to as the ``Down-In" design. We also examine the ``Down-Out" design, where the dimension of $\beta_{\text{cal}}$ is equal to the out-channels of $W_{\rm down}$, i.e., $d^\prime$. In this scenario, the dimension of the cross modality factor $f_C$ remains unchanged, but the projection matrix $M_S$ is reconfigured from $d^m\times d$ to $d^m\times d^\prime$ to align the out-channel dimensions. Beyond comparing ``In" and ``Out", we also explore the calibration of \texttt{Upsample} weights with CMT, leading to a comparison of four variants. As indicated in \cref{tab:pos}, calibrating the in-channels of the downsampling matrix yields the best performance.
    \item \textbf{Optimal settings.} Further ablations are conducted to identify the optimal CMT settings, including the encoder layers where CMT is applied, and the dimension of the modality factor $f_C \in \mathbb{R}^{d_m}$. The results are shown in~\cref{tab:layer} and \cref{tab:size} respectively. Considering both R@1 and R@Sum, CMT achieves optimal results when applied to the final 2 layers with a factor dimension of 32. We hypothesize that applying CMT in higher layers is more effective, as feature spaces of the two modalities converge more closely, whereas earlier application might negatively impact the training process.
\end{itemize}

\noindent {\bf Results on More Datasets.} In the main paper, we present ablations on MSR-VTT due to space limit. \cref{tab:other_dts} shows ablation results on other datasets. These new results lead to similar conclusions: equipping the model with temporal adaptation (TA) consistently improves performance across all datasets, confirming the importance of TA in our tasks; Additionally, by facilitating modality alignment, CMT significantly enhances the performance of each model. Since CMT incurs negligible extra parameters (less than 0.1\% of vanilla CLIP), it can be conveniently used to boost model capabilities. \\
\begin{table}[h]
\centering
\small
  \begin{tabular}{l|cc}
    \toprule
    Designs & T2V & V2T  \\
    \midrule
    Down-In  & 46.2/\textbf{202.1} & \textbf{47.2/205.9} \\
    Down-Out  & \textbf{46.8}/201.8 & 47.0/204.9 \\
    Up-In  & 46.5/200.7 & 46.3/204.5 \\
    Up-Out  & 46.0/200.5 & 46.5/203.7 \\
    \bottomrule
  \end{tabular}
  \vspace{-1mm}
  \caption{Ablations on the design of CMT on MSR-VTT \cite{msrvtt} using the CLIP (ViT-B/16) backbone~\cite{CLIP}. We set factor dimension to 32, and use CMT in the last 2 layers of encoders by default. }
  \label{tab:pos}
\end{table}
\begin{table}[t]
\centering
\small
\begin{tabular}{l|cc}
    \toprule
    Layers & T2V & V2T  \\
    \midrule
    No CMT & 45.9/200.9 & 46.6/204.4 \\
    Last 1  & \textbf{46.6}/201.3 & 46.8/204.1 \\
    Last 2  & 46.2/\textbf{202.1} & \textbf{47.2/205.9} \\
    Last 3  & 45.8/201.0 & 46.6/205.8 \\
    Last 6  & 45.9/201.0 & 46.8/204.9 \\
    Last 12  & 45.4/199.7 & 46.5/203.9 \\
    \bottomrule
  \end{tabular}
  \vspace{-1mm}
  \caption{Ablations on layers using CMT on MSR-VTT \cite{msrvtt} using the  CLIP (ViT-B/16) backbone~\cite{CLIP} with factor dimension 32. ``Last n'' refers to using CMT in the last n layers of the visual/text encoders. }
  \label{tab:layer}
\end{table}
\begin{table}[t]
\centering
\small
  \begin{tabular}{l|cc}
    \toprule
    Dim & T2V & V2T  \\
    \midrule
    8 & \textbf{46.7}/202.0 & 46.5/204.7 \\
    16 &  46.4/202.1 & 46.5/205.1 \\
    32 &  46.2/202.1 & \textbf{47.2/205.9} \\
    64 &  46.3/200.6 & 46.0/204.1 \\
    \bottomrule
  \end{tabular}
  \vspace{-1mm}
  \caption{Ablations on the factor dimension in CMT on MSR-VTT \cite{msrvtt}.
  The backbone used is CLIP (ViT-B/16) \cite{CLIP}. \\}
  \label{tab:size}
\end{table}

\begin{table*}
\renewcommand{\arraystretch}{1.2}
  \scriptsize
  \centering
  \begin{tabular}{l|cccc|cccc||cccc|cccc}
  \toprule
  & \multicolumn{8}{c||}{\textbf{MSVD}} & \multicolumn{8}{c}{\textbf{LSMDC}}\\
     \hline
    \multirow{2}*{Settings} & \multicolumn{4}{c|}{Text-to-Video} & \multicolumn{4}{c||}{Video-to-Text} & \multicolumn{4}{c|}{Text-to-Video} & \multicolumn{4}{c}{Video-to-Text}  \\
     & R@1 & R@5 & R@10 & Sum & R@1 & R@5 & R@10 & Sum & R@1 & R@5 & R@10 & Sum & R@1 & R@5 & R@10 & Sum \\
     \hline
     \textbf{Ours} & \textbf{49.4} & \textbf{78.3} & \textbf{87.0} & \textbf{214.8} & \textbf{71.8} & \textbf{93.0} & \textbf{96.4} & \textbf{261.2} & 23.2 & \textbf{43.9} & 53.2 & \textbf{120.3} & \textbf{24.0} & \textbf{42.8} & 52.1 & \textbf{118.8} \\
     w/o TA & 49.3 & 78.2 & \textbf{87.0} & 214.5 & 70.8 & 93.2 & 96.1 & 260.2 & 23.1 & 42.7 & 52.4 & 118.2 & 23.6 & 42.2 & \textbf{52.4} & 118.3 \\
     w/o CMT & 49.0 & \textbf{78.3} & 86.9 & 214.2 & 71.0 & 92.3 & 96.2 & 259.4 & \textbf{23.4} & 43.2 & \textbf{53.3} & 119.9 & 23.0 & 41.6 & 51.8 & 116.5 \\
     \hline
     \hline
 & \multicolumn{8}{c||}{\textbf{Didemo}} & \multicolumn{8}{c}{\textbf{ActivityNet}}\\
     \hline
    \multirow{2}*{Settings} & \multicolumn{4}{c|}{Text-to-Video} & \multicolumn{4}{c||}{Video-to-Text} & \multicolumn{4}{c|}{Text-to-Video} & \multicolumn{4}{c}{Video-to-Text}  \\
     & R@1 & R@5 & R@10 & Sum & R@1 & R@5 & R@10 & Sum & R@1 & R@5 & R@10 & Sum & R@1 & R@5 & R@10 & Sum \\
     \hline
     \textbf{Ours} & \textbf{44.3} & \textbf{72.1} & \textbf{80.5} & \textbf{196.8} & {42.7} & 73.0 & \textbf{81.9} & \textbf{197.6} & 42.7 & 74.2 & \textbf{85.8} & 202.7 & \textbf{44.0} & 74.4 & 86.0 & 204.4 \\
     w/o TA & 43.5 & 71.9 & 80.4 & 195.8 & \textbf{43.2} & 72.2 & 81.2 & 196.6 & 42.1 & 72.9 & 84.5 & 199.6 & 42.9 & 73.4 & 85.5 & 201.8 \\
     w/o CMT & 43.8 & 71.8 & 80.4 & 195.9 & 42.4 & \textbf{73.2} & 81.5 & 197.1 & \textbf{42.9} & \textbf{74.5} & 85.7 & \textbf{203.1} & 43.6 & \textbf{75.0} & \textbf{86.5} & \textbf{205.2} \\
     \bottomrule
  \end{tabular}
  \caption{Ablation of TA and CMT modules by removing one at a time from MV-Adapter. ``Sum'' represents the sum +of R@1/5/10 in Text-to-Video or Video-to-Text task. The backbone is CLIP (ViT-B/16) \cite{CLIP}. }
\label{tab:other_dts}
\end{table*}
\section{Case Study}
We summarize two failure modes of our method from comprehensive case studies: 
\begin{itemize}
    \item Since we use a fixed number of frames per video in training, our method may fail to capture fine movements in long videos as the differences between frames aggregate.
    \item Results may be incorrect when the caption is related to the audio contents.
\end{itemize} Examples of these two modes are put into \texttt{long\_video} and \texttt{audio} directories respectively. These directories have been zipped together with this document and uploaded to \href{https://github.com/zhangbw17/MV-Adapter}{https://github.com/zhangbw17/MV-Adapter}. Each directory contains one retrieval result consisting of caption.txt (the query used to search), gt\_*.mp4, and pred\_*.mp4 (the ground truth and predicted video, where * represents the index number of videos in MSR-VTT  \cite{msrvtt}.)

For the sake of clarity, we present a detailed analysis of each example below.

\textbf{long\_video\_0.} In this case, the clips featuring ``people" are quite concentrated and ``fade" quickly. Given the video's duration of 27 seconds, the long intervals between extracted frames result in key information being omitted, making it impossible to match the description. Therefore, another video, where the fading effect and the characters are clearer, is returned instead.

\textbf{long\_video\_1.}  The groundtruth video is long and the shot that corresponds to the target in the query (walking down a short runway) is relatively short. Since the number of input frames is fixed, non-target information in videos tends to dominate the input, making the model fail to parse out the fine movement (``walking" and ``short runway") from the groundtruth. Eventually, the model returns a similar video that contains ``walking" but on a ``long runway" (should be a short runway).

\textbf{audio\_0.} Audio information is necessary in order to determine the topic of talking.

\textbf{audio\_1.} Though the retrieved result is visually similar to groundtruth, the contents of the talk do not match that of the query text. With the help of audio content (like transcripts from ASR), the results can be corrected.
\section{Efficiency Analysis}
\label{sub_sec:efficiency}MV-Adapter has three types of newly added parameters: down-sampling, a lightweight transformer and up-sampling, the parameter complexity of which are $O(d\times d^{\prime})$, $O(d^{\prime}\times d)$, and $O( (d^{\prime})^2)$, respectively. Temporal calibration's parameter complexity is also $O( (d^{\prime})^2)$. CMT only introduces $O(d^m + d^m\times d)$ parameters. As a result, MV-Adapter is rather parameter-efficient as $d^m,d^\prime\ll d$. The total increase in parameters compared with the CLIP backbone is about 2.4\% when $d$ is 768 and $d^\prime$ is 64. Due to its small number of tunable parameters, MV-Adapter is highly parameter-efficient in both deployment and training stages.